# Enhancing the retrieval performance by combing the texture and edge features

Mohamed Eisa
Computer Science Department
PortSaid University
Mmmeisa@yahoo.com

Amira Eletrebi
Computer Science Department
Mansoura University
Amera_eletrebe@yahoo.com

Ebrahim Elhenawy
Computer Science Department
Zagazig University
Henawy2000@yahoo.com

*Abstract*— In this paper, anew algorithm which is based on geometrical moments and local binary patterns (LBP) for content based image retrieval (CBIR) is proposed. In geometrical moments, each vector is compared with the all other vectors for edge map generation. The same concept is utilized at LBP calculation which is generating nine LBP patterns from a given $3 \times 3$ pattern. Finally, nine LBP histograms are calculated which are used as a feature vector for image retrieval. Moments are important features used in recognition of different types of images. Two experiments have been carried out for proving the worth of our algorithm. The results after being investigated shows a significant improvement in terms of their evaluation measures as compared to LBP and other existing transform domain techniques.

Keywords- CBIR; Feature extraction; geometrical moments; Local Binary Patterns.

## I. INTRODUCTION

With the rapid expansion of worldwide network and advances in information technology there is an explosive growth of multimedia databases and digital libraries. This demands an effective tool that allow users to search and browse efficiently through such a large collections [1].

In many areas of commerce, government, academia, hospitals, entertainment, and crime preventions large collections of digital images are being created. Usually, the only way of searching these collections was by using keyword indexing, or simply by browsing. However, as the databases grew larger, people realized that the traditional keywords based methods to retrieve a particular image in such a large collection are inefficient. To describe the images with keywords with a satisfying degree of concreteness and detail, we need a very large and sophisticated keyword system containing typically several hundreds of different keywords. One of the serious drawbacks of this approach is the need of trained personnel not only to attach keywords to each image (which may take several minutes for one single image) but also to retrieve images by selecting keywords, as we usually need to know all keywords to choose good ones. Further, such a keyword based approach is mostly influenced by subjective decision about image content and also it is very difficult to change a keyword based system afterwards. Therefore, new techniques are needed to overcome these limitations [2].

Digital image databases however, open the way to content based searching. It is common phrase that an image speaks thousands of words. So instead of manual annotation by text based keywords, images should be indexed by their own visual contents, such as color, texture and shape. [3]The main advantage of this method is its ability to support the visual queries. Hence researchers turned attention to content based image retrieval (CBIR) methods.

Several methods achieving effective feature extraction have been proposed in the literature [4].

[5] Introduced the histogram intersection distance metric to measure the distance between the histograms of images. Stricker et al [6] used the first three central moments called mean, standard deviation and skewness of each color for image retrieval. Pass et al. introduced color coherence vector (CCV) [7].

The recently proposed local binary pattern (LBP) features are designed for texture description. The recently,[8], proposed the LBP and these LBPs are converted to rotational invariant for texture classification. we proposed the rotational invariant texture classification using feature distributions. [9] used the LBP operator facial expression analysis and recognition. Heikkila et al. proposed the background modeling and detection by using LBP. Huang et al. proposed the extended LBP for shape localization. Heikkila et al. used the LBP for interest region description. Li et al. used the combination of Gabor filter and LBP for texture segmentation. Zhang et al. proposed the local derivative pattern for face recognition [10]. They have considered LBP as a nondirectional first order local pattern, which are the binary results of the first-order derivative in images.

## II. GEOMETRICAL MOMENTS

The shape of an object is a very important character in human's perception, recognition, and comprehension. Because geometric shape represents the essential characteristic of an object, and has invariance with respect to translation, scale and orientation, the analysis and discernment like geometry are of important significance in computer vision. Historically, Hu published the first significant paper on the use of image moment invariants for two-dimensional pattern recognition applications [11]. His approach is based on the work of the 19th century mathematicians Boole, Cayley and Sylvester, and on the theory of algebraic forms.

$$M_1 = \mu_{20} + \mu_{02} \tag{1}$$

$$M_2 = (\mu_{20} - \mu_{02})^2 + 4\mu_{11}^2 \tag{2}$$

$$M_3 = (\mu_{30} - 3\mu_{12})^2 + 3(\mu_{21} + \mu_{03})^2 \tag{3}$$

$$M_4 = (\mu_{30} + \mu_{12})^2 + (\mu_{21} + \mu_{03})^2 \tag{4}$$

$$\begin{aligned}M_5 =& (\mu_{30}-3\mu_{12})(\mu_{30}+\mu_{12})[(\mu_{30}+\mu_{12})^2 - 3(\mu_{21}+\mu_{03})^2] \\ &+ (3\mu_{21}-\mu_{03})(\mu_{21}+\mu_{03})[3(\mu_{30}+\mu_{12})^2 - (\mu_{21}+\mu_{03})^2]\end{aligned} \tag{5}$$

$$\begin{aligned}M_6 =& (\mu_{20} - \mu_{02})[(\mu_{30}+\mu_{12})^2 - (\mu_{21}+\mu_{03})^2] \\ &+ 4\mu_{11}(\mu_{30}+\mu_{12})(\mu_{21}+\mu_{03})\end{aligned} \tag{6}$$

$$\begin{aligned}M_7 =& (3\mu_{21}-\mu_{03})(\mu_{30}+\mu_{21})[(\mu_{31}+\mu_{12})^2 - 3(\mu_{21}+\mu_{03})^2] \\ &+ (3\mu_{12}-\mu_{30})(\mu_{21}+\mu_{31})[3(\mu_{30}+\mu_{12})^2 - (\mu_{21}+\mu_{03})^2]\end{aligned} \tag{7}$$

Geometric moments of a 1D signal $S(x)$ are defined by [12]:

$$M_n(x) = \int_{-\omega}^{\omega} S(x+t)t^n dt \qquad n = 0,1,2,... \tag{8}$$

Where $M_n(x)$ is the moment of order n calculated from a window of size $(2\omega+1)$ pixels centered at the point $x$. Geometric moments of a 2D image $I(x, y)$ are defined by [13]:

$$M_{mn}(x,y) = \int_{-\varpi_1}^{\varpi_1}\int_{-\varpi_2}^{\varpi_2} I(x+u, y+v)u^m v^n du dv \quad m,n=0,1,2.. \tag{9}$$

Where $M_{m,n}(x)$ is the moment of order $(m, n)$ calculated from a window of size $(2\varpi_1+1) \times (2\varpi_2+1)$ pixels centered at the pixel $(x, y)$.

### III. LOCAL BINARY PATTERNS

[14] proposed the local binary pattern (LBP) operator which describes the surroundings of a pixel by generating a bit-code from the binary derivatives of a pixel as a complementary measure for local image contrast. The LBP operator takes the eight neighboring pixels using the center gray value as a threshold. The operator generates a binary code 1 if the neighbor is greater or equal than the center otherwise generates a binary code 0. The eight neighboring binary code can be represented by a 8-bit number [15]. The LBP operator outputs for all the pixels in the image can be accumulated to form a histogram. Fig.1 shows an example of LBP operator. For given a center pixel in the image, LBP value is computed by comparing it with those of its neighborhoods:

$$LBP_{P,R} = \sum_{i=0}^{p-1} 2^i x f(g_i - g_c) \tag{10}$$

$$f(x) = \begin{cases} 1 & x \geq 0 \\ 0 & else \end{cases} \tag{11}$$

Where $g_c$ is the gray value of the center pixel, $g_i$ is the gray value of its neighbors, P is the number of neighbors and R is the radius of the neighborhood. Fig. 2 shows the examples of circular neighbor sets for different configurations of (P,R).

The LBP measure the local structure by assigning unique identifiers, the binary number, to various microstructures in the image. Thus [16], LBP capture many structures in one unified framework. In the example in Fig. 3(b), the local structure is a vertical edge with a leftward intensity gradient. Other microstructures are assigned different LBP codes, e.g., corners and spots, as illustrated in Fig. 4. By varying the radius R and the number of samples P, the structures are measured at different scales, and LBP allows for measuring large scale structures without smoothing effects, as is, e.g., the case for Gaussian-based filters.

Example | | | Binary Pattern | | |
---|---|---|---|---|---
6 | 5 | 2 | 1 | 0 | 2
7 | 6 | 1 | 1 |   | 1
9 | 8 | 7 | 1 | 0 | 7

Weights | | | LBP value | | |
---|---|---|---|---|---
8 | 4 | 2 |   |   |  
16 |   | 1 |   | 248 |  
32 | 64 | 128 |   |   |  

LBP=8+16+32+64+128=248

Fig. 1: LBP calculation for $3 \times 3$ pattern

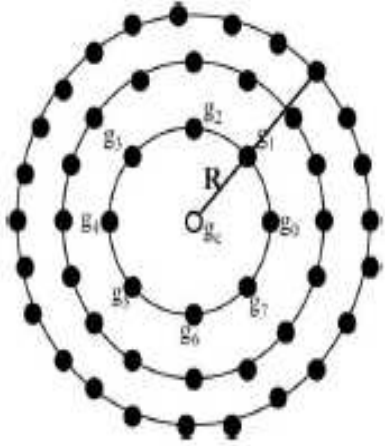

Fig. 2: Circular neighborhood sets for different (P,R)

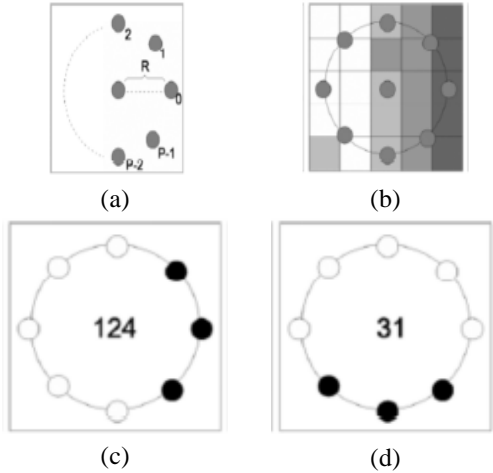

(a)  (b)

(c)  (d)

Fig. 3. Illustration of LBP. (a)The LBP filter is defined by two parameters; the circle radius R and the number of samples P on the circle. (b) Local structure is measured w. r. t. a given pixel by placing the center of the circle in the position of that pixel. (c) Samples on the circle are binarized by thresholding with the intensity in the center pixel as threshold value. Black is zero and white is one. The example image shown in (b) has an LBP code of 124. (d) Rotating the example image in (b) 900 clockwise reduces the LBP code to 31, which is the smallest possible code for this binary pattern. This principle is used to achieve rotation invariance.

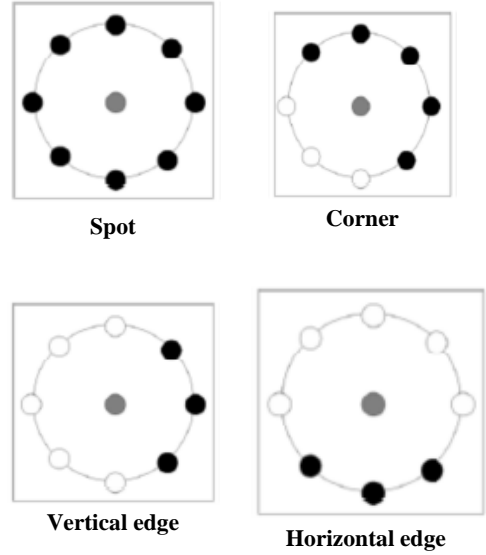

**Spot**  **Corner**

**Vertical edge**  **Horizontal edge**

Fig 4: Various microstructures measured by LBP. The gray circle indicates the center pixel. Black and white circles are binarized samples; black is zero and white is one.

IV. FEATURES EXTRACTION

The weighted graph $(L_i \ X_i \ et \ al.,)$ with no self loops is $G = (V, E, W)$, where $V = \{1, 2, ......., N\}$ the node set is ($N = m.n$ is the total number of pixels in $Q \in R^{mxn}$) $E \subseteq VxV$ represents the edge set, and $W = (w_{ij})_{NxN}$ denotes an affinity matrix with the element $w_{ij}$ being the edge weight between nodes i and j.

Based on the geometric moment's theory we compare the each pixel of $3x3$ pattern with remaining eight pixel gray values for generating binary code [17]. Finally, nine LBP patterns are collected for LBP histogram calculation and these are used as a feature vector for image retrieval [18].

A. *Proposed System Framework (GMLBP)@*

Algorithm:

Input: Image;          Output: Retrieval Result
1. Load the input image.
2. Collect the $3x3$ pattern for a center pixel i.
    · Construct the graph cut for $3x3$ pattern.
    · Generate nine LBP patterns.
    · Go to next center pixel.
3. Calculate the geometric moments LBP (GMLBP) histograms [19].
4. Form the feature vector by concatenating the nine LBP features [20].
5. Calculate the best matches using Eq. (12) [21].
6. Retrieve the number of top matches [22].

*B. Similarity Measurement*

In the presented work $d_1$ similarity distance metric is used as shown below:

$$D(Q, I_1) = \sum_{i=1}^{L_g} \left| \frac{f_{L,i} - f_{Q,i}}{1 + f_{L,i} + f_{Q,i}} \right| \quad (12)$$

Where Q is query image, $L_g$ is feature vector length, $I_1$ is image in database; $f_{1,i}$ is $i^{th}$ feature of image $I$ in the database, $f_{Q,i}$ is $i^{th}$ feature of query image Q.

## V. EXPERIMENTS AND EVALUATIONS

When the input data is too large to be processed and redundant, then the input data will be transformed into a reduced representation set of features. Transforming the input data into the set of features is called features extraction [23]. Features extraction involves simplifying the amount of resources required to describe a large set of data accurately [24]. When performing analysis of complex data one of the major problems is the number of variables involved [25].

Here,

$$\Pr ecision(P) = \frac{No.\,of\ Relevant\ Images\ Retrieved}{Total\ No.of\ Image\ Retrieved} \times 100 \quad (13)$$

$$Group\ \Pr ecision\ (GP) = \frac{1}{N_1} \sum_{i=1}^{N_1} P \quad (14)$$

$$Average\ Retrieval\ \Pr ecision(ARR) = \frac{1}{\Gamma_1} \sum_{j=1}^{\Gamma_1} GP \quad (15)$$

$$Recall(R) = \frac{Number\ of\ relevant\ image\ retrikeved}{Total\ Number\ of\ relevant\ images} \quad (16)$$

$$Group\ Recall\ (GR) = \frac{1}{N_1} \sum_{i=1}^{N_1} R \quad (17)$$

$$Average\ Retrieval\ Rate\ (ARR) = \frac{1}{\Gamma_1} \sum_{j=1}^{\Gamma_1} GR \quad (18)$$

Where $N_1$ is number of relevant images and $\Gamma_1$ is number of groups.

**Table 1**: Retrieval results of proposed method (GM) and LBP in terms of average retrieval precision (ARP) (%)

| Method | Number of top matches considered | | | | | | | | |
|---|---|---|---|---|---|---|---|---|---|
| | 1 | 3 | 5 | 7 | 9 | 11 | 13 | 15 | 16 |
| LBP | 100 | 89.1 | 84.6 | 81.7 | 79.01 | 76.33 | 73.86 | 71.1 | 69.6 |
| GM | 100 | 93.1 | 89.7 | 87.2 | 85.02 | 82.71 | 80.47 | 77.8 | 76.4 |

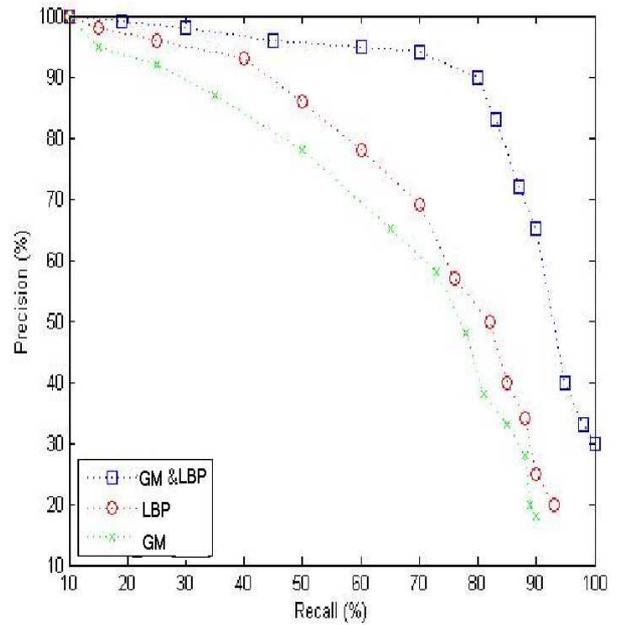

Fig. 5. Average retrieval performance

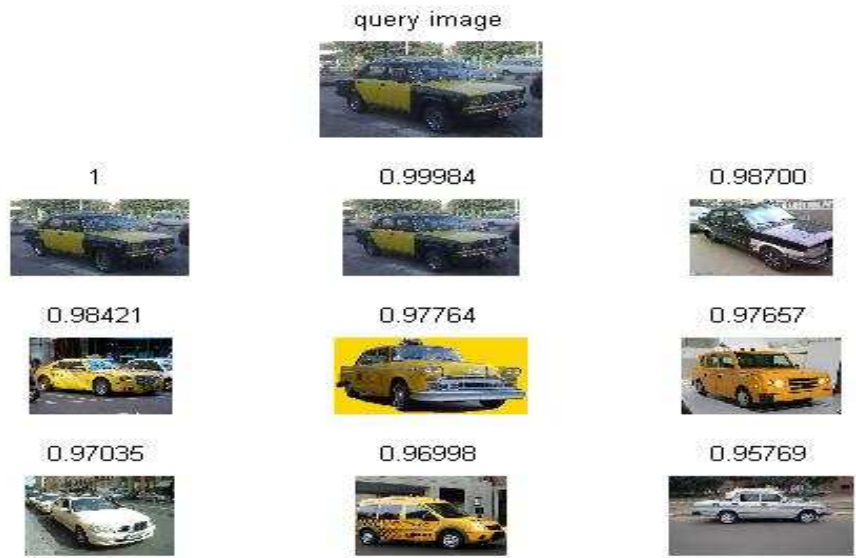

Fig .6: Retrieval results using geometrical moments

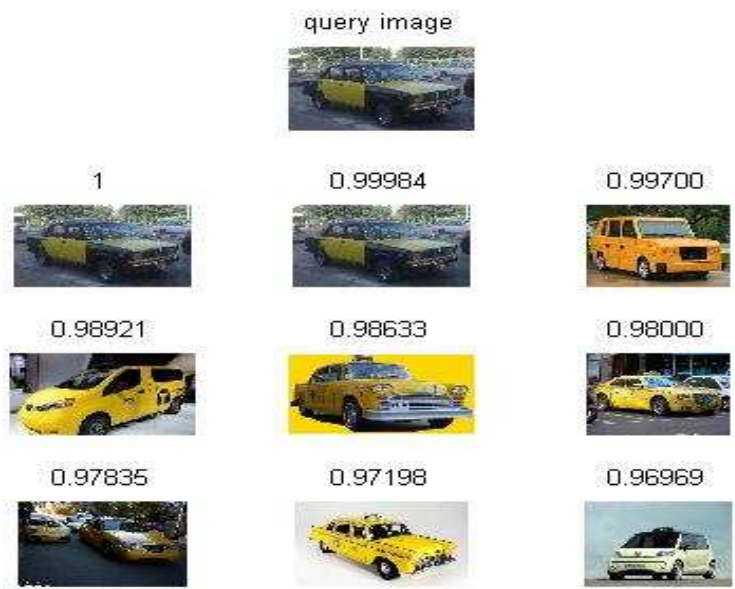

Fig. 7: Retrieval results using LBP

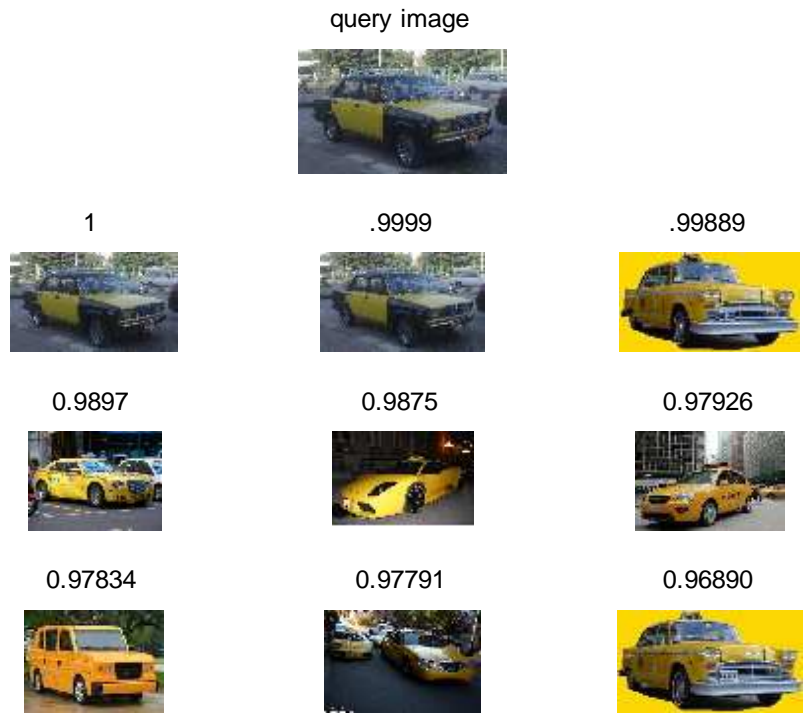

Fig 8: Retrieval results using combining geometrical moments & LBP

## VI. CONCLUSIONS

A new algorithm which is based on the geometrical moments and local binary patterns (LBP) for content based image retrieval (CBIR) is proposed in this paper. The proposed method extracts the nine LBP patterns from a given $3\times 3$ pattern and these are used as the features. Two experiments have been carried out for proving the worth of our algorithm. The results after being investigated shows a significant improvement in terms of their evaluation measures as compared to LBP and other existing transform domain techniques.